\def\year{2018}\relax
\documentclass[letterpaper]{article} 
\usepackage{aaai18}  
\usepackage{times}  
\usepackage{helvet}  
\usepackage{courier}  
\usepackage{url}  
\usepackage{graphicx}  

\usepackage{algorithm}
\usepackage{algorithmic}

\begin{document}
%
\title{LSTD: A Low-Shot Transfer Detector for Object Detection}
\author{Hao Chen\thanks{Equally-contributed}$^{1,2}$, Yali Wang$^{*1}$, Guoyou Wang$^{2}$, Yu Qiao$^{1, 3}$\\
$^{1}$
Shenzhen Institutes of Advanced Technology, Chinese Academy of Sciences, China\\
$^{2}$Huazhong University of Science and Technology, China\\
$^{3}$The Chinese University of Hong Kong, Hong Kong\\
\{hao$_{-}$chen, gywang\}@hust.edu.cn, \{yl.wang, yu.qiao\}@siat.ac.cn\\}
\maketitle

\begin{abstract}
Recent advances in object detection are mainly driven by deep learning with large-scale detection benchmarks.
However,
the fully-annotated training set is often limited for a target detection task,
which may deteriorate the performance of deep detectors.
To address this challenge,
we propose a novel low-shot transfer detector (LSTD) in this paper,
where we leverage rich source-domain knowledge to construct an effective target-domain detector with very few training examples.
The main contributions are described as follows.
First,
we design a flexible deep architecture of LSTD to alleviate transfer difficulties in low-shot detection.
This architecture can integrate the advantages of both SSD and Faster RCNN in a unified deep framework.
Second,
we introduce a novel regularized transfer learning framework for low-shot detection,
where
the transfer knowledge (TK) and background depression (BD) regularizations are proposed to leverage object knowledge respectively from source and target domains,
in order to further enhance fine-tuning with a few target images.
Finally,
we examine our LSTD on a number of challenging low-shot detection experiments,
where
LSTD outperforms other state-of-the-art approaches.
The results demonstrate that LSTD is a preferable deep detector for low-shot scenarios.
\end{abstract}

\begin{figure*}[t]
\centering
\includegraphics[width= 0.9\textwidth]{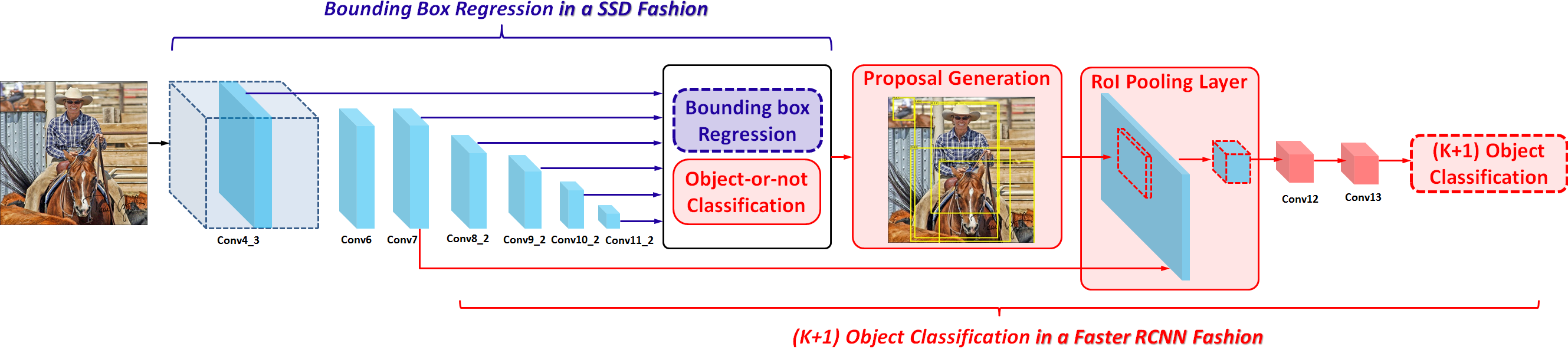}
\caption{Basic deep architecture of low-shot transfer detector (LSTD).
Since we aims at detecting objects effectively in the low-shot setting,
we integrate the core designs of both SSD \cite{Liueccv2016} and Faster RCNN \cite{Renpami2016} in a non-trivial manner,
i.e.,
the multi-convolutional-layer design for bounding box regression and the coarse-to-fine design for object classification.
Both designs in LSTD are crucial when training with few examples.
More details can be found in Section \ref{Basic Deep Architecture of LSTD}.}
\label{LSTD}
\end{figure*}

\section{Introduction}
Over the past years,
a number of deep learning approaches have achieved remarkable performance in object detection \cite{Girshick2014,Girshick2016Fast,ren2015faster,redmon2016you,Liueccv2016,He2017}.
However,
the successes of these deep detectors heavily depend on the large-scale detection benchmarks with fully-annotated bounding boxes.
In practice,
the fully-annotated training set may be limited for a given target detection task,
which can restrict the power of deep detectors.

One popular solution is to collect extra detection images but with easily-annotated labels (e.g., image-level supervision).
In this case,
weakly-supervised \cite{diba2016weakly,kantorov2016contextlocnet,bilen2016weaklyddn,li2016weakly,cinbis2017,TangCVPR2017}
or
semi-supervised approaches \cite{Tang2016,Singh2016,Liang2016,dong2017few}
can be used to relieve annotation difficulties in detection.
However,
the performance of these detectors is often limited,
because of lacking sufficient supervision on the training images.

An alternative solution is to perform transfer learning on deep models \cite{yosinski2014nips},
due to its success on image classification \cite{Donahue2014,Razavian2014}.
Compared to the weakly/semi-supervised solution,
this is often a preferable choice without extra data collection.
More importantly,
the source-domain knowledge is an effective supervision to generalize the learning procedure of target domain,
when the training set is scarce.
However,
transfer learning for low-shot detection is still an open challenge due to the following reasons.
First,
it is inappropriate to apply the general transfer strategy of object detection (i.e., initializing deep detectors from the pretrained deep classifiers),
when the target detection set is limited.
This is mainly because,
fine-tuning with such small target sets is often hard to eliminate the task difference between detection and classification.
Second,
deep detectors are more prone to overfitting during transfer learning,
compared to deep classifiers.
It is mainly because that,
detectors have to learn more object-specific representations for both localization and classification tasks of detection.
Finally,
simple fine-tuning may reduce transferability,
since it often ignores the important object knowledge from both source and target domains.

To address the challenges above,
we propose a low-shot transfer detector (LSTD) in this paper,
which is the first transfer learning solution for low-shot detection,
according to our best knowledge.
The main contributions are described as follows.
\textbf{First},
we design a novel deep architecture of LSTD,
which can boost low-shot detection via incorporating the strengths of two well-known deep detectors,
i.e., SSD \cite{Liueccv2016} and Faster RCNN \cite{Renpami2016},
into a non-trivial deep framework.
In addition,
our LSTD can flexibly perform bounding box regression and object classification on two different model parts,
which promotes a handy transfer learning procedure for low-shot detection.
\textbf{Second},
we propose a novel regularized transfer learning framework for LSTD,
where we can flexibly transfer from source-domain LSTD to target-domain LSTD,
avoiding the task difference (such as transferring from classification to detection in the general strategy).
In this case,
target-domain LSTD can sufficiently integrate common object characteristics of large-scale detection data from source-domain LSTD.
Furthermore,
we enhance fine-tuning with a novel regularization consisting of transfer knowledge (TK) and background depression (BD).
TK transfers the source-object-label knowledge for each target-domain proposal,
in order to generalize low-shot learning in the target domain.
BD integrates the bounding box knowledge of target images as an extra supervision on feature maps,
so that LSTD can suppress background disturbances while focus on objects during transferring.
\textbf{Finally},
our LSTD outperforms other state-of-the-art approaches on a number of low-shot detection experiments,
showing that LSTD is a preferable deep detector for low-shot scenarios.

\begin{figure*}[t]
\centering
\includegraphics[width= 0.9 \textwidth]{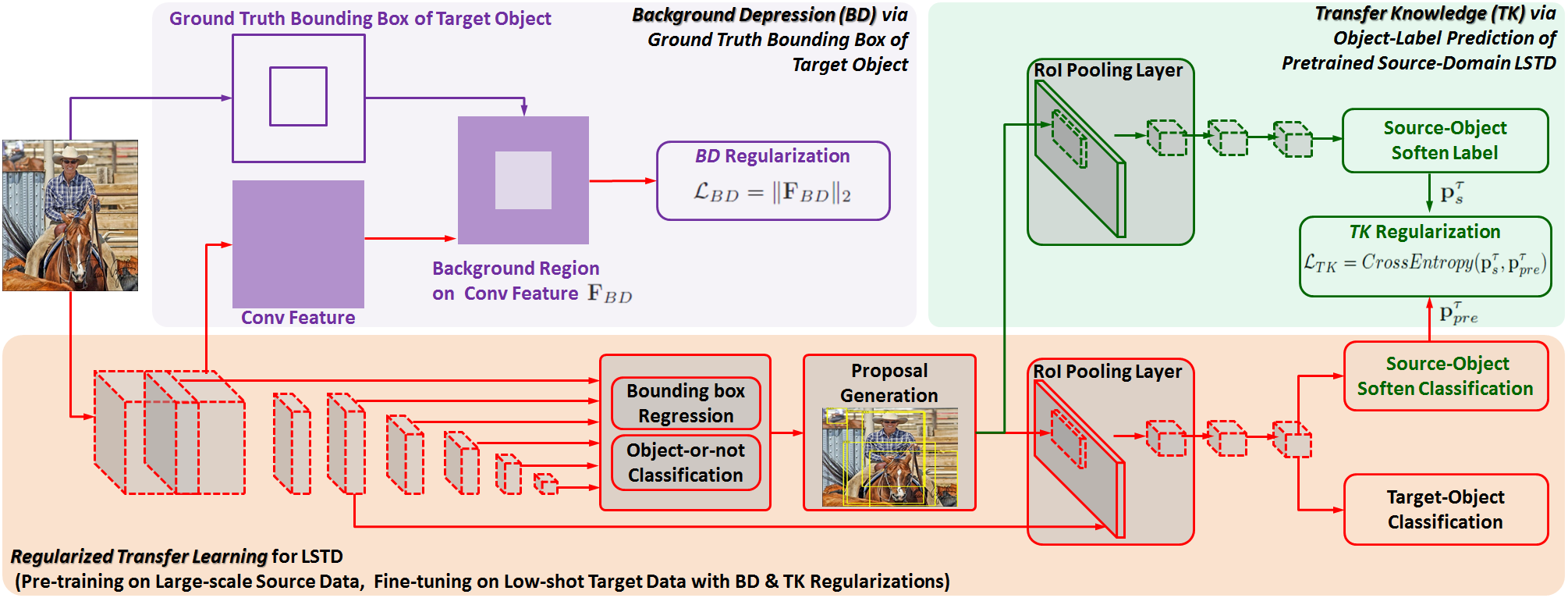}
\caption{Regularized transfer learning for LSTD.
First,
we train the source-domain LSTD with a large-scale source data set.
Second,
we initialize the target-domain LSTD using the pretrained source-domain LSTD.
Finally,
we use the small-scale target data to fine-tune the target-domain LSTD with the proposed low-shot detection regularization
(i.e., background depression and transfer-knowledge).
More details can be found in Section \ref{Transfer Learning with Low-Shot Detection Regularization}.}
\label{Regularization}
\end{figure*}

\section{Related Works}

\textbf{Object Detection}.
Recent developments in object detection are driven by deep learning models \cite{Girshick2014,Girshick2016Fast,ren2015faster,redmon2016you,Liueccv2016,He2017}.
We mainly discuss two popular deep detectors,
i.e.,
Faster RCNN \cite{ren2015faster} and SSD \cite{Liueccv2016},
which are closely relevant to our approach.
Faster RCNN is a popular region-proposal architecture,
where object proposals are firstly generated from region proposal network (RPN) and then fed into Fast RCNN \cite{Girshick2016Fast} for end-to-end detection.
SSD is a widely-used one-stage detection architecture,
where the multi-layer design of bounding box regression can efficiently localize objects with various sizes.
Both approaches have achieved the stat-of-the-art detection performance on the large-scale data sets (e.g. Pascal VOC and COCO).
However,
one may be stuck in troubles by directly using Faster RCNN and SSD with a few training examples,
since the architecture design in these approaches lacks the low-shot considerations.
For Faster RCNN,
the bounding box regression design is not effective for low-shot detection.
The main reason is that the bounding box regressor in Faster RCNN is separate for each object category.
In this case,
the regressor for target domain has to be randomly initialized without any pre-trained parameters from source domain.
Obviously,
it can deteriorate model transferability,
especially for low-shot target cases.
For SSD,
the object classification design is not effective for low-shot detection.
It is mainly because,
SSD directly addresses the $(K+1)$-object classification (i.e., $K$ objects + background) on the default boxes,
without any coarse-to-fine analysis.
When few images are available,
the random initialization of this classifier may trap into training difficulties.

\textbf{Low-shot Learning}.
Low-shot learning is mainly inspired by the fact that humans can learn new concepts with little supervision \cite{Lake2015}.
Recently,
a number of related approaches have been proposed by
Bayesian program learning  \cite{Lake2015},
memory machines \cite{Graves2014,Santoro2016,Vinyals2016},
and so on.
However,
the existing low-shot approaches are mainly designed for the standard classification task \cite{Xu2017,Hariharan2017}.
For object detection,
a low-shot detector \cite{dong2017few} has been proposed recently in a semi-supervised learning framework.
But like other semi-supervised \cite{Hoffman2014,Tang2016,Singh2016,Liang2016} or weakly-supervised \cite{diba2016weakly,kantorov2016contextlocnet,bilen2016weaklyddn,li2016weakly,cinbis2017} detectors,
its performance is often limited because of lacking effective supervision on the training images.
Finally,
transfer learning \cite{yosinski2014nips,Razavian2014} is a reasonable choice when the training set is small,
since large-scale source benchmarks can generalize the learning procedure in the low-shot target domain \cite{Fei2006,Hinton2015}.
However,
simple fine-tuning with standard deep detectors may reduce the detection performance,
as both object localization and classification in these architectures lack effective transfer learning designs for low-shot detection.
Furthermore,
the object knowledge from both source and target domains may not be fully considered when fine-tuning with a few target images.

Different from the previous approaches,
we propose a novel low-shot transfer detector (LSTD) for object detection with few annotated images.
Specifically,
we first introduce a flexible deep architecture of LSTD,
where
we leverage the advantages from both Faster RCNN and SSD to alleviate transfer difficulties in low-shot learning.
Furthermore,
we design a regularized transfer learning framework for LSTD.
With the proposed regularization,
LSTD can integrate the object knowledge from both source and target domains to enhance transfer learning of low-shot detection.

\section{Low-Shot Transfer Detector (LSTD)}
In this section,
we describe the proposed low-shot transfer detector (LSTD) in detail.
First,
we introduce the basic deep architecture of our LSTD,
and explain why it is an effective detector when the target data set is limited.
Then,
we design a novel regularized transfer learning framework for LSTD,
where
the background-depression and transfer-knowledge regularizations can enhance low-shot detection,
by leveraging object knowledge respectively from both target and source domains.

\subsection{Basic Deep Architecture of LSTD}
\label{Basic Deep Architecture of LSTD}

To achieve the low-shot detection effectively,
we first need to alleviate the training difficulties in the detector,
when a few training images are available.
For this reason,
we propose a novel deep detection architecture in Fig. \ref{LSTD},
which can take advantage of two state-of-the-art deep detectors,
i.e.,
SSD \cite{Liueccv2016} and Faster RCNN \cite{Renpami2016},
to design effective bounding box regression and object classification for low-shot detection.

First,
we design \textbf{bounding box regression in the fashion of SSD}.
Specifically,
for each of selected convolutional layers,
there are a number of default candidate boxes (over different ratios and scales) at every spatial location of convolutional feature map.
For any candidate box matched with a ground truth object,
the regression loss (smooth L1) is used to penalize the offsets (box centers, width and height) error between the predicted and ground truth bounding boxes.
As a result,
\textit{this multiple-convolutional-feature design in SSD is suitable to localize objects with various sizes.
This can be especially important for low-shot detection,
where we lack training samples with size diversity.}
More importantly,
the regressor in SSD is shared among all object categories,
instead of being specific for each category as in Faster RCNN.
In this case,
\textit{the regression parameters of SSD, which are pretrained on the large-scale source domain, can be re-used as initialization in the different low-shot target domain.}
This avoids re-initializing bounding box regression randomly,
and thus reduces the fine-tuning burdens with only a few images in the target domain.

Second,
we design \textbf{object classification in the fashion of Faster RCNN}.
Specifically,
we first address the binary classification task for each default box,
to check if a box belongs to an object or not.
According to the classification score of each box,
we choose object proposals of region proposal network (RPN) in Faster RCNN.
Next,
we apply the region-of-interests (ROI) pooling layer on a middle-level convolutional layer,
which produces a fixed-size convolutional feature cube for each proposal.
Finally,
instead of using the fully-connected layers in the original Faster RCNN,
we use two convolutional layers on top of ROI pooling layer for $(K+1)$-object classification.
This further reduces overfitting with fewer training parameters.
Additionally,
\textit{the coarse-to-fine classifier may be more effective to alleviate training difficulties of transfer learning,
compared to the direct $(K+1)$-object classification for each default box in SSD}.
Our key insight is that,
objects in source and target may share some common traits (e.g., clear edge, uniform texture),
compared with background.
Hence,
we propose to transfer this knowledge with the object-or-not classifier,
which helps to generate better target-object proposals and thus boost the final performance.
On the contrary,
the direct $(K+1)$ classifier has to deal with thousands of randomly-selected proposals.


\textbf{Summary}.
Our deep architecture aims at reducing transfer learning difficulties in low-shot detection.
To achieve it,
we flexibly leverage the core designs of both SSD and Faster RCNN in a non-trivial manner,
i.e.,
the multi-convolutional-layer design for bounding box regression and the coarse-to-fine design for object classification.
Additionally,
our LSTD performs bounding box regression and object classification on two relatively separate places,
which can further decompose the learning difficulties in low-shot detection.

\subsection{Regularized Transfer Learning for LSTD}
\label{Transfer Learning with Low-Shot Detection Regularization}

After designing a flexible deep architecture of LSTD,
we introduce an end-to-end regularized transfer learning framework for low-shot detection.
The whole training procedure is shown in Fig. \ref{Regularization}.
\textbf{First},
we train LSTD in the source domain,
where
we apply a large-scale source data set to train LSTD in Fig. \ref{LSTD}.
\textbf{Second},
we fine-tune the pre-trained LSTD in the target domain,
where a novel regularization is proposed to further enhance detection with only a few training images.
Specifically,
the total loss of fine-tuning can be written as
\begin{equation}
\mathcal{L}_{total}=\mathcal{L}_{main}+\mathcal{L}_{reg},
\label{eq:totalloss}
\end{equation}
where
\textit{the main loss} $\mathcal{L}_{main}$ refers to the loss summation of multi-layer bounding box regression and coarse-to-fine object classification in LSTD.
Note that,
the object categories between source and target can be relevant but different,
since low-shot detection aims at detecting the previously-unseen categories from little target data.
In this case,
the $(K+1)$-object classification (i.e., $K$ objects + background) has to be randomly re-initialized in the target domain,
even though the bounding box regression and object-or-not classification can be initialized from the pre-trained LSTD in the source domain.
Consequently,
fine-tuning with only $\mathcal{L}_{main}$ may still suffer from the unsatisfactory overfitting.
To further enhance low-shot detection in the target domain,
we design \textit{a novel regularization} $\mathcal{L}_{reg}$,
\begin{equation}
\mathcal{L}_{reg}=\lambda_{BD}\mathcal{L}_{BD}+\lambda_{TK}\mathcal{L}_{TK},
\label{eq:regloss}
\end{equation}
where
$\mathcal{L}_{BD}$ and $\mathcal{L}_{TK}$ respectively denote the background-depression and transfer-knowledge terms,
$\lambda_{BD}$ and $\lambda_{TK}$ are the coefficients for $\mathcal{L}_{BD}$ and $\mathcal{L}_{TK}$.

\begin{figure}[t]
\centering
\includegraphics[width= 0.3 \textwidth]{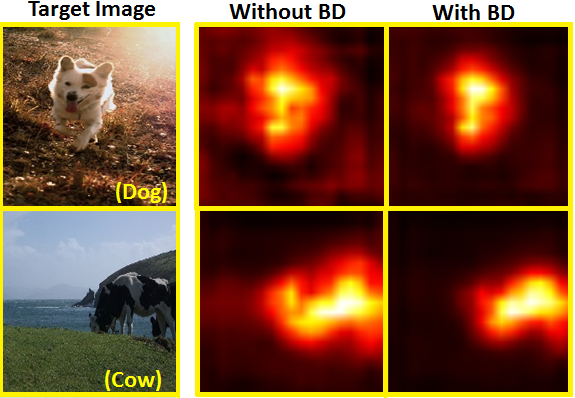}
\caption{Background-Depression (BD) regularization.
The feature heatmap is obtained by averaging the convolutional feature cube (conv5$_{-}$3) over feature channels.
BD can successfully alleviate background disturbances on the feature heatmap,
and thus allow LSTD to focus on target objects.}
\label{BDRegularization}
\end{figure}

\begin{figure}[t]
\centering
\includegraphics[width= 0.35 \textwidth]{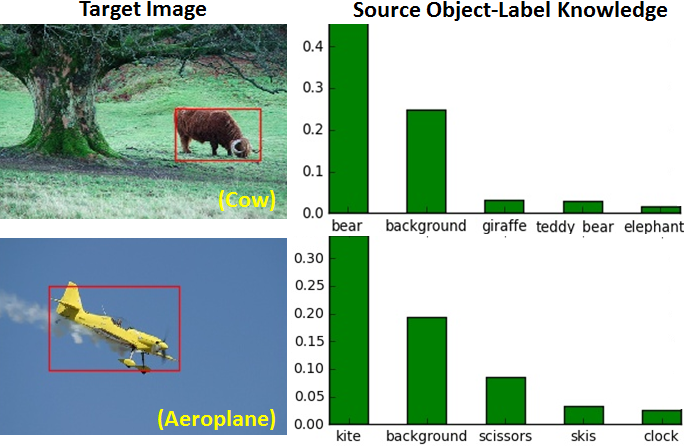}
\caption{Transfer-Knowledge (TK) regularization.
For a target-object proposal (red box: the proposal with the highest score),
we plot the top 5 soften-labels of source-object via Eq. (\ref{Eq:sk}).
TK can effectively provide important source-domain knowledge for target object proposals,
i.e.,
the target object \textit{Cow} (or \textit{Aeroplane}) is strongly relevant to \textit{Bear} (or \textit{Kite}) in source,
due to color (or shape) similarity.}
\label{TKRegularization}
\end{figure}

\begin{table*}[t]
\centering
\begin{tabular}{l|l|l}
  \hline
  Tasks & Source (large-scale training set) & Target (1/2/5/10/30 training images per class) \\
  \hline
  Task 1 & COCO (standard 80 classes, 118,287 training images) &  ImageNet2015 (chosen 50 classes) \\
  Task 2& COCO (chosen 60 classes, 98,459 training images)   &  VOC2007 (standard 20 classes) \\
  Task 3& ImageNet2015 (chosen 181 classes, 151,603 training images) & VOC2010 (standard 20 classes) \\
  \hline
\end{tabular}
\caption{Data description.
The object categories for source and target are carefully selected to be non-overlapped,
in order to evaluate if LSTD can detect unseen object categories from few training shots in the target domain.}
\label{DataDescription}
\end{table*}

\textbf{Background-Depression (BD) Regularization}.
In the proposed deep architecture of LSTD,
bounding box regression is developed with the multi-convolutional-layer design of SSD.
Even though this design can reduce the training difficulties for objects with various sizes,
the complex background may still disturb the localization performance in the low-shot scenario.
For this reason,
we propose a novel background-depression (BD) regularization,
by using object knowledge in the target domain (i.e., ground-truth bounding boxes in the training images).
Specifically,
for a training image in the target domain,
we first generate the convolutional feature cube from a middle-level convolutional layer of LSTD.
Then,
we mask this convolutional cube with the ground-truth bounding boxes of all the objects in the image.
Consequently,
we can identify the feature regions that are corresponding to image background,
namely $\mathbf{F}_{BD}$.
To depress the background disturbances,
we use L2 regularization to penalize the activation of $\mathbf{F}_{BD}$,
\begin{equation}
\mathcal{L}_{BD}=\|\mathbf{F}_{BD}\|_{2}.
\label{eq:BDregularization}
\end{equation}
With this $\mathcal{L}_{BD}$,
\textit{LSTD can suppress background regions while pay more attention to target objects,
which is especially important for training with a few training images}.
It is clearly shown in Fig. \ref{BDRegularization} that our BD regularization can help LSTD to reduce the background disturbances.

\textbf{Transfer-Knowledge (TK) Regularization}.
The coarse-to-fine classification of LSTD can alleviate the difficulties in object classification,
since we can use the pretrained object-or-not classifier in the target domain.
However,
the $(K+1)$-object classifier has to be randomly re-initialized for $K$ new objects (plus background) in the target domain,
due to the category difference between source and target.
In this case,
simply fine-tuning this classifier with target data may not make full use of source-domain knowledge.
As shown in Fig. \ref{TKRegularization},
the target object \textit{Cow} (or \textit{Aeroplane}) is strongly relevant to the source-domain category \textit{Bear} (or \textit{Kite}),
due to color (or shape) similarity.
For this reason,
we propose an novel transfer-knowledge (TK) regularization,
where
the object-label prediction of source network is used as source-domain knowledge to regularize the training of target network for low-shot detection.
Note that,
object classification in the detection task requires to be applied for each object proposal,
instead of the entire image in the standard image classification task.
Hence,
we design TK regularization for each object proposal in the target domain.

\textbf{(I) Source-Domain Knowledge}.
First,
we feed a training image respectively to source-domain and target-domain LSTDs.
Then,
we apply target-domain proposals into the ROI pooling layer of source-domain LSTD,
which can finally generate a knowledge vector from the source-domain object classifier,
\begin{equation}
\mathbf{p}_{s}^{\tau}=Softmax(\mathbf{a}_{s}/\tau),
\label{Eq:sk}
\end{equation}
where
$\mathbf{a}_{s}$ is the pre-softmax activation vector for each object proposal,
$\tau>1$ is a temperature parameter that can produce the soften label with richer label-relation information \cite{Hinton2015}.

\textbf{(II) Target-Domain Prediction of Source-Domain Categories}.
To incorporate the source-domain knowledge $\mathbf{p}_{s}^{\tau}$ into the training procedure of target-domain LSTD,
we next modify the target-domain LSTD into a multi-task learning framework.
Specifically,
we add a source-object soften classifier at the end of target-domain LSTD.
For each target proposal,
this classifier produces a soften prediction of source object categories,
\begin{equation}
\mathbf{p}_{pre}^{\tau}=Softmax(\mathbf{a}_{pre}/\tau),
\label{Eq:sp}
\end{equation}
where
$\mathbf{a}_{pre}$ is the pre-softmax activation for each proposal.

\textbf{(III) TK Regularization}.
With the knowledge $\mathbf{p}_{s}^{\tau}$ of source-domain LSTD and the soften prediction $\mathbf{p}_{pre}^{\tau}$ of target-domain LSTD,
we apply the cross entropy loss as a TK regularization,
\begin{equation}
\mathcal{L}_{TK}=CrossEntropy(\mathbf{p}_{s}^{\tau}, \mathbf{p}_{pre}^{\tau}).
\label{Eq:TKLoss}
\end{equation}
In this case,
\textit{the source-domain knowledge can be integrated into the training procedure of target domain,
which generalizes LSTD for low-shot detection in the target domain}.

\begin{algorithm}[t]
\begin{algorithmic}
   \STATE \textbf{1. Pre-training on Large-scale Source Domain}
   \STATE $\bullet$ Source-domain LSTD is trained with a large-scale detection benchmark.
   \STATE \textbf{2. Initialization for Small-scale Target Domain}
   \STATE $\bullet$ The last layer of $(K+1)$-object classifier is randomly initialized, due to object difference in source and target.
   \STATE $\bullet$ All other parts of target-domain LSTD are initialized from source-domain LSTD.
   \STATE \textbf{3. Fine-tuning for Small-scale Target Domain}
   \STATE $\bullet$ We fine-tune target-domain LSTD with BD and TK regularizations (Eq. \ref{eq:totalloss}-\ref{Eq:TKLoss}),
          based on the small training set.
\end{algorithmic}
\caption{Regularized Transfer Learning of LSTD}
\label{alg}
\end{algorithm}

\textbf{Summary}.
To reduce overfitting with few training images,
we propose an end-to-end regularized transfer learning framework for LSTD.
According to our best knowledge,
it is the first transfer learning solution for low-shot detection.
The whole training procedure is shown in Alg. \ref{alg},
where
we sufficiently leverage the pretrained source-domain LSTD to generalize the target-domain LSTD.
Furthermore,
we design a novel regularization (i.e., BD and TK) to effectively enhance fine-tuning with limited target training set.

\begin{figure*}[t]
\centering
\includegraphics[width=0.85\textwidth]{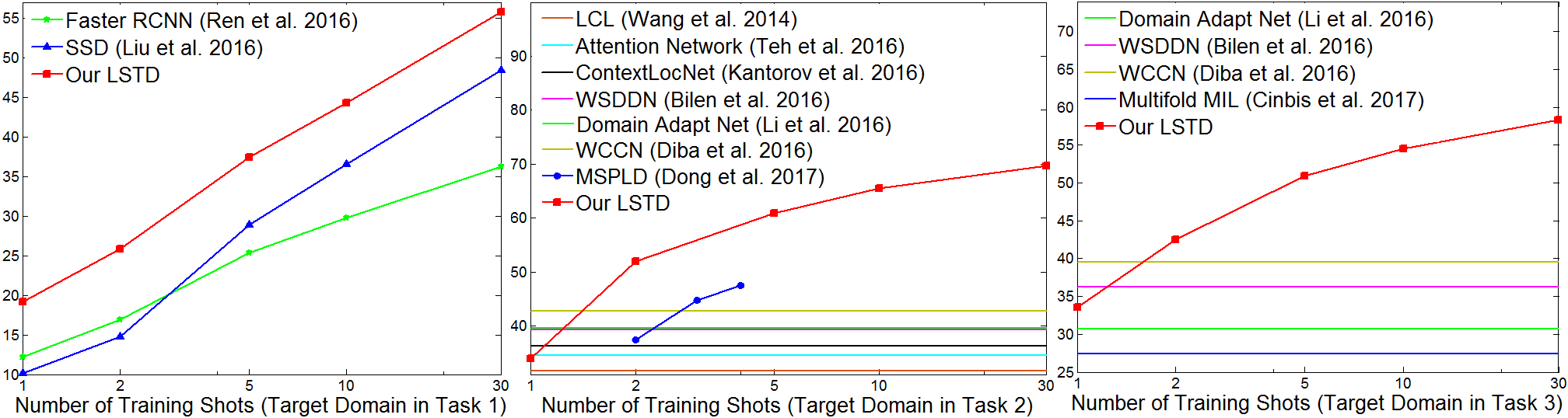}
\caption{Comparison (mAP) with the state-of-the-art.
Note that,
all the training images are used in weakly-supervised detectors.
Hence,
all these approaches are flat lines in the plots.
More explanations can be found in the text.}
\label{CompareWithStateOfTheArt}
\end{figure*}

\section{Experiment}
In this section,
we conduct a number of challenging low-shot detection experiments to show the effectiveness of LSTD.

\textbf{Data Sets}.
Since our LSTD is a low-shot detector within a regularized transfer learning framework,
we adopt a number of detection benchmarks,
i.e.,
COCO \cite{coco},
ImageNet2015 \cite{imagenet},
VOC2007 and VOC2010 \cite{voc},
respectively as source and target of three transfer tasks (Table \ref{DataDescription}).
The training set is large-scale in the source domain of each task,
while
it is low-shot in the target domain (1/2/5/10/30 training images for each target-object class).
Furthermore,
the object categories for source and target are carefully selected to be non-overlapped,
in order to evaluate if our LSTD can detect unseen object categories from few training shots in the target domain.
Finally,
we use the standard PASCAL VOC detection rule on the test sets to report mean average precision (mAP) with 0.5 intersection-over-union (IOU).
Note that,
the target domain of task 1 is ImageNet2015 with the chosen 50 object classes.
Hence,
we define a test set for this target domain,
where we randomly sample 100 images in each target-object class of ImageNet2015.
To be fair,
the training and test images in this target domain are non-overlapped.
The target domains of task 2 and 3 refer to the standard VOC2007 and VOC2010.
Hence,
we use the standard test sets for evaluation.

\textbf{Implementation Details}.
Unless stated otherwise,
we perform our LSTD as follows.
\textbf{First},
the basic deep architecture of LSTD is build upon VGG16 \cite{simonyan2014very},
similar to SSD and Faster RCNN.
For bounding box regression,
we use the same structure in the standard SSD.
For object classification,
we apply the ROI pooling layer on conv7,
and add two convolutional layers (conv12: $3\times3\times256$, conv13: $3\times3\times256$ for task 1/2/3) before the $(K+1)$-object classifier.
\textbf{Second},
we train LSTD in a regularized transfer learning framework (Alg. \ref{alg}).
In the source domain,
we feed 32 training images into LSTD for each mini-batch in task 1/2/3,
and train bounding box regressor and object-or-not binary classifier in the fashion of SSD \cite{Liueccv2016}.
Subsequently,
100/100/64 proposals (after non-maximum suppression of top 1000 proposals at 0.65) are selected to train the (K+1)-object classifier.
In the target domain,
all the training settings are the same as the ones in the source domain,
except that
64/64/64 proposals are selected to train the (K+1)-object classifier,
the background depression regularization is used on conv5$_{-}$3,
the temperature parameter in the transfer-knowledge regularization is 2 as suggested in \cite{Hinton2015}.
and the weight coefficients for both background depression and transfer-knowledge are 0.5.
\textbf{Finally},
the optimization strategy for both source and target is Adam \cite{Kingma2015},
where
the initial learning rate is 0.0002 (with 0.1 decay),
the momentum/momentum2 is 0.9/0.99,
and the weight decay is 0.0001.
All our experiments are performed on Caffe \cite{jia2014caffe}.

\begin{table}[t]
\centering
\begin{tabular}{l|cc}
  \hline
  Deep Models & Large Source & Low-shot Target\\
  \hline
  Faster RCNN & 21.9  & 12.2  \\
  SSD & 25.1 & 10.1 \\
  \hline
  Our LSTD$_{conv5_{-}3}$ & 24.7  & 15.9  \\
  Our LSTD$_{conv7}$ & \textbf{25.2}  & \textbf{16.5}  \\
  \hline
\end{tabular}
\caption{Comparison with Faster RCNN and SSD for Task1.
We compare LSTD with its closely-related SSD \cite{Liueccv2016} and Faster RCNN \cite{Renpami2016} in both source and target domains.
The mAP results show that,
LSTD is a more effective deep architecture for low-shot detection in target.
Furthermore,
the structure of LSTD itself is robust,
with regards to different convolutional layers for ROI pooling.}
\label{LSTDSourceTarget}
\end{table}

\begin{table}[t]
\centering
\begin{tabular}{l|ccccc}
\hline
Shots for Task 1 & 1 & 2 & 5 & 10 & 30  \\
\hline
LSTD$_{FT}$ &  16.5 & 21.9 & 34.3 & 41.5 & 52.6 \\
LSTD$_{FT + TK}$ & 18.1 & 25.0 & 35.6 & 43.3 & 55.0 \\
LSTD$_{FT + TK + BD}$ & \textbf{19.2} & \textbf{25.8} & \textbf{37.4} & \textbf{44.3} & \textbf{55.8} \\
\hline
Shots for Task 2 & 1 & 2 & 5 & 10 & 30  \\
\hline
LSTD$_{FT}$& 27.1 & 46.1 & 57.9 & 63.2 & 67.2 \\
LSTD$_{FT + TK}$ & 31.8 & 50.7 & 60.4 & 65.1 & 69.0 \\
LSTD$_{FT + TK + BD}$ & \textbf{34.0} & \textbf{51.9} & \textbf{60.9} & \textbf{65.5} & \textbf{69.7} \\
\hline
Shots for Task 3 & 1 & 2 & 5 & 10 & 30  \\
\hline
LSTD$_{FT}$& 29.3 & 37.2  & 48.1 & 52.1 & 56.4 \\
LSTD$_{FT + TK}$ & 32.7 & 40.8 & 49.7 & 54.1 & 57.9 \\
LSTD$_{FT + TK + BD}$ & \textbf{33.6} & \textbf{42.5} & \textbf{50.9} & \textbf{54.5} & \textbf{58.3} \\
\hline
\end{tabular}
\caption{Regularized transfer learning for LSTD.
$FT$: standard fine-tuning.
$TK$: transfer knowledge (TK) regularization.
$BD$: background depression (BD) regularization.
The mAP results show that,
our low-shot detection regularization ($TK+BD$) can significantly help the fine-tuning procedure of LSTD,
when the training set is scarce in target.}
\label{LSTDTransfer}
\end{table}

\begin{table}[t]
\centering
\begin{tabular}{l|ccccc}
\hline
Tasks & $BD_{conv5_{-}3}$ & $BD_{conv7}$  \\
\hline
Task1 & \textbf{19.2}  & 18.9 \\
Task2 & 34.0  & \textbf{34.5}  \\
Task3 & \textbf{33.6}  & 33.4 \\
\hline
\end{tabular}
\caption{Background Depression (BD) regularization.
We perform LSTD for one-shot detection in the target domain,
where
BD regularization is implemented on different convolutional layers when fine-tuning.
The mAP results show that BD is robust to different convolutional layers.}
\label{BDdiffconv}
\end{table}

\begin{figure*}
\centering
\includegraphics[width=0.83\textwidth]{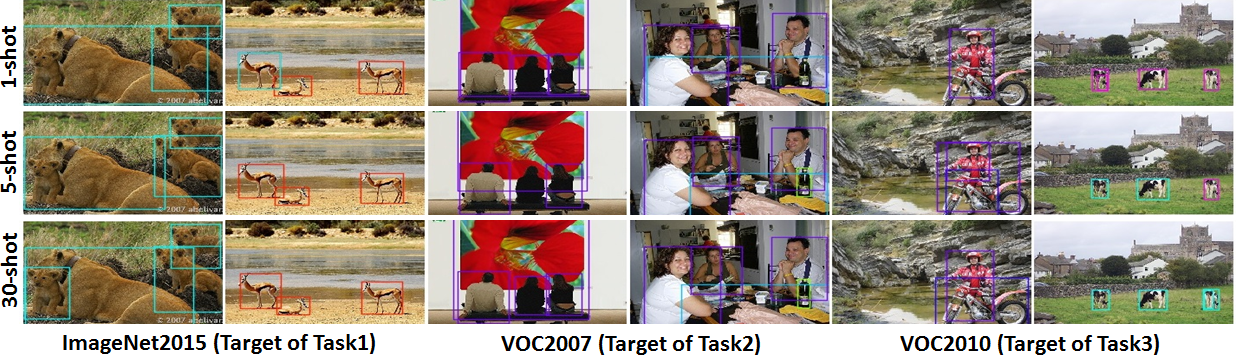}
\caption{Detection visualization.
Our LSTD can successfully detect most target objects with only a few training shots (such as one or five shots) in the target domain,
illustrating that it is an effective and robust deep approach for low-shot detection.}
\label{detectdifferentshot}
\end{figure*}

\subsection{Properties of LSTD}
To investigate the properties of LSTD,
we evaluate the effectiveness of its key designs.
To be fair,
when we explore different settings for one design,
all other designs are with the basic setting in the implementation details.

\textbf{Basic Deep Structure of LSTD}.
We first evaluate the basic deep structure of LSTD respectively in the source and target domains,
where we compare it with the closely-related SSD \cite{Liueccv2016} and Faster RCNN \cite{Renpami2016}.
For fairness,
we choose task 1 to show the effectiveness.
The main reason is that,
the source data in task 1 is the standard COCO detection set,
where SSD and Faster RCNN are well-trained with the state-of-the-art performance.
Hence,
we use the published SSD \cite{Liueccv2016} and Faster RCNN \cite{Renpami2016} in this experiment,
where
the size of input images for SSD and our LSTD is $300\times300$,
and Faster RCNN follows the settings in the original paper.
In Table \ref{LSTDSourceTarget},
we report mAP on the test sets of both source and target domains in task 1.
One can see that,
our LSTD achieves a competitive mAP in the source domain.
It illustrates that LSTD can be a state-of-art deep detector for large-scale training sets.
More importantly,
our LSTD outperforms both SSD and Faster RCNN significantly for low-shot detection in the target domain (one training image per target category),
where
all approaches are simply fine-tuned from their pre-trained models in the source domain.
It shows that,
LSTD yields a more effective deep architecture for low-shot detection,
compared to SSD and Faster RCNN.
This can also be found in Fig. \ref{CompareWithStateOfTheArt},
when we change the number of training shots in the target domain.
Finally,
we investigate the structure robustness in LSTD itself.
As the bounding box regression follows the standard SSD,
we explore the (K+1)-object classifier in which we choose different convolutional layers (conv$5_{-}3$ or conv$7$) for ROI pooling.
The results are comparable in Table \ref{LSTDSourceTarget},
showing the architecture robustness of LSTD.
For consistency,
we use conv$7$ for ROI pooling in all our experiments.

\textbf{Regularized Transfer Learning for LSTD}.
We mainly evaluate if the proposed regularization can enhance transfer learning for LSTD,
in order to boost low-shot detection.
As shown in Table \ref{LSTDTransfer},
our background-depression (BD) and transfer knowledge (TK) regularizations can significantly improve the baseline (i.e., fine-tuning),
especially when the training set is scarce in the target domain (such as one-shot).
Additionally,
we show the architecture robustness of BD regularization in Table \ref{BDdiffconv}.
Specifically,
we perform LSTD for one-shot detection in the target domain,
where
BD regularization is implemented on different convolutional layers for fine-tuning.
One can see that BD is generally robust to different convolutional layers.
Hence,
we apply BD on conv$5_{-}3$ in all our experiments for consistency.

\subsection{Comparison with the State-of-the-art}
We compare our LSTD to the recent state-of-the-art detection approaches,
according to mAP on the test set of target domain.
The results are shown in Fig. \ref{CompareWithStateOfTheArt}.
First,
LSTD outperforms SSD \cite{Liueccv2016} and Faster RCNN \cite{Renpami2016},
when changing the number of training images in Task 1.
It shows the architecture superiority of LSTD for low-shot detection.
Second,
LSTD outperforms other weakly-supervised \cite{wang2014weakly,teh2016attention,kantorov2016contextlocnet,bilen2016weaklyddn,li2016weakly,diba2016weakly,cinbis2017} and
semi-supervised \cite{dong2017few} detectors,
when the number of training shots is beyond two in Task 2 and 3.
Note that,
we pick the results of these weakly/semi-supervised detectors from the original papers,
and compare them with our LSTD on the same test set.
It shows that our LSTD is more effective and efficient,
as our LSTD only requires a few fully-annotated training images in the target domain. 
On the contrary,
both weakly-supervised and semi-supervised approaches require the full training set
(i.e.,
weakly-supervised: all training images with only image-level labels,
semi-supervised: a few fully-annotated training shots + other training images with only image-level labels).
In fact,
our LSTD outperforms these detectors on task 3 with only 0.4\% training data,
and can be competitive to fully-supervised detectors (LSTD:69.7, SSD:68.0, Faster RCNN:69.9) with only 11\% of training set.

\subsection{Visualization}

In this section,
we qualitatively visualize our LSTD.
\textbf{First},
we visualize the detection results in Fig. \ref{detectdifferentshot},
based on various numbers of training shots in the target domain.
As expected,
1-shot LSTD may localize but misclassify some objects (e.g., cows),
due to the scarce training set.
But this misclassification can be largely clarified with only 5-shot.
It illustrates that,
our LSTD is an effective and robust deep approach for low-shot detection.
\textbf{Second},
we briefly analyze the error mode of LSTD on VOC 2007.
For 2-shot LSTD,
the error for Animal comes from (23\%Loc,71\%Sim, 4\%BG),
where the notations of (Loc, Sim, BG) follow \cite{Girshick2014}.
As expected,
the main error in low-shot detection may come from confusion with similar objects.

\section{Conclusion}
In this paper,
we propose a novel low-shot transfer detector (LSTD) to address object detection with a few training images.
First,
we design a flexible deep architecture of LSTD to reduce transfer difficulties of low-shot detection.
Second,
we train LSTD within a regularized transfer learning framework,
where we introduce a novel low-shot detection regularization (i.e., TK and BD terms) to generalize fine-tuning with a few target images.
Finally,
our LSTD outperforms other state-of-the-art approaches on a number of challenging experiments,
demonstrating that LSTD is a preferable deep detector for low-shot scenarios.

\textbf{Acknowledgments}.
This work was supported in part by National Key Research and Development Program of China (2016YFC1400704),
National Natural Science Foundation of China (U1613211, 61633021, 61502470),
and Shenzhen Research Program (JCYJ20160229193541167, JCYJ20150925163005055).

\bibliography{reference}
\bibliographystyle{aaai}

\end{document}